\documentclass[DIV=10]{scrartcl}

\usepackage{graphicx}
\usepackage{amsmath}
\usepackage{amssymb}
\usepackage{amsmath, amssymb}
\usepackage{newtxtext, newtxmath}

\DeclareFontFamily{T1}{qcr}{\hyphenchar\font-1}
\DeclareFontShape{T1}{qcr}{m}{n}{<-> s*[0.85]qx-qcrr}{}
\DeclareFontShape{T1}{qcr}{m}{it}{<-> s*[0.85]qx-qcrri}{}
\DeclareFontShape{T1}{qcr}{m}{sl}{<-> ssub* qcr/m/it}{}
\DeclareFontShape{T1}{qcr}{bx}{n}{<-> s*[0.85]qx-qcrb}{}
\DeclareFontShape{T1}{qcr}{bx}{it}{<-> s*[0.85]qx-qcrbi}{}
\DeclareFontShape{T1}{qcr}{bx}{sl}{<-> ssub* qcr/bx/it}{}

\usepackage{booktabs}
\usepackage{makecell}
\usepackage[absolute]{textpos}
\usepackage{nicefrac}
\usepackage{enumitem}
\usepackage{placeins}
\usepackage{float}
\usepackage[backend=biber,style=ieee,doi=false]{biblatex}
\usepackage{multirow}
\usepackage{microtype}
\usepackage{expl3}
\usepackage{svg}
\usepackage[acronym]{glossaries}

\usepackage{xcolor}

\newglossaryentry{gop}
{
  name={GOP},
  description={Group of Pictures},
  first={\glsentrydesc{gop} (\glsentrytext{gop})},
  plural={GOPs},
  descriptionplural={Groups of Pictures},
  firstplural={\glsentrydescplural{gop} (\glsentryplural{gop})}
}
\newacronym{auc}{AUC}{Area Under the Receiver Operating Characteristic}
\newacronym{roc}{ROC}{Receiver Operating Characteristic}
\newacronym{sa}{SA}{Self Attention}
\newacronym{msa}{MSA}{Multihead Self Attention}
\newacronym{vit}{ViT}{Vision Transformer}
\newacronym{prnu}{PRNU}{Photo Response Non-uniformity}
\newacronym{vcdi}{VDI}{Video Device Identification}
\newacronym{vcdv}{VDM}{Video Device Matching}
\newacronym{qp}{QP}{Quantization Parameter}
\newacronym{cnn}{CNN}{Convolution Neural Network}
\newacronym{tpr}{TPR}{True Positive Rate}
\newacronym{tnr}{TNR}{True Negative Rate}

\newcommand{\G}{\mathcal{G}}

\ExplSyntaxOn
\clist_new:N \g_doc_bold_letters_clist
\clist_set:Nn \g_doc_bold_letters_clist {q,k,v,r,t,A,Z,U,V,W}

\clist_map_inline:Nn \g_doc_bold_letters_clist {
    \cs_gset:cpn {b#1} {
        \boldsymbol{#1}
    }
}
\ExplSyntaxOff

\renewcommand{\bt}{\boldsymbol{o}}

\newcommand{\smallsp}{\hspace*{0.5pt}}
\newcommand{\dt}{D_{\bt}}
\newcommand{\dr}{D_{\br}}

\newcommand{\sa}{\operatorname{SA}}
\newcommand{\msa}{\operatorname{MSA}}

\newcommand{\appname}{H4VDM}
\newcommand{\appfullname}{H.264 Video Device Matching (H4VDM)}

\newcommand{\etal}{\emph{et al.}}

\usepackage{hyperref}

\usepackage[capitalize,noabbrev]{cleveref}

\hypersetup{
pdftitle={H4VDM: H.264 Video Device Matching},
pdfauthor={Ziyue Xiang, Paolo Bestagini, Stefano Tubaro, Edward J. Delp},
pdfkeywords={H.264 Video Compression, Video Device Matching, Digital Video Forensics, Deep Learning},
}

\begin{document}

\title{H4VDM: H.264 Video Device Matching}

\renewcommand*{\glslinkcheckfirsthyperhook}{\setkeys{glslink}{hyper=false}}

\newcounter{dummycounter}
\setcounter{dummycounter}{1}

\newcommand{\syma}{\|}
\newcommand{\symd}{\S}
\newcommand{\symb}{$^\dagger$}
\newcommand{\symc}{$^\ddagger$}

\author{
\small Ziyue Xiang\symb, Paolo Bestagini\symc, Stefano Tubaro\symc, Edward J. Delp\symb\\
\footnotesize \symb Video and Image Processing Lab (VIPER), School of Electrical and Computer Engineering,\vspace*{-0.5em}\\
\footnotesize Purdue University, West Lafayette, Indiana, USA \vspace*{0.3em}\\
\footnotesize \symc Dipartimento di Elettronica, Informazione e Bioingegneria, Politecnico di Milano, Milano, Italy
}
\date{}
\maketitle

%\let\oldfootnote\thefootnote
%\renewcommand*{\thefootnote}{\fnsymbol{footnote}}
%\footnotetext[1]{Corresponding author}
%\let\thefootnote\oldfootnote

\begin{abstract}
Methods that can determine if two given video sequences are captured by the same device (e.g., mobile telephone or digital camera) can be used in many forensics tasks. 
In this paper we refer to this as ``video device matching''.
In open-set video forensics scenarios it is easier to determine if two video sequences were captured with the same device than identifying the specific device.
In this paper, we propose a technique for open-set video device matching.
Given two H.264 compressed video sequences, our method can determine if they are captured by the same device, even if our method has never encountered the device in training. We denote our proposed technique as H.264 Video Device Matching (H4VDM).
H4VDM uses H.264 compression information extracted from video sequences to make decisions.
It is more robust against artifacts that alter camera sensor fingerprints, 
and it can be used to analyze relatively small fragments of the H.264 sequence.
We trained and tested our method on a publicly available video forensics dataset consisting of 35 devices, where our proposed method demonstrated good performance.

\vspace*{1em}

\end{abstract}

\section{Introduction}\label{sec::introduction}

\glsreset{vcdi}
\glsreset{vcdv}

\gls{vcdi} is one of the most important tasks in multimedia forensics \cite{kurosawa1999ccd,lukas2006digital,bhagtani2022overview}.
A \gls{vcdi} method can associate a video with a specific source device (e.g., a specific camera).
\gls{vcdi} is valuable in forensics investigations and court defense.

A large amount of \gls{vcdi} techniques rely on the analysis of video camera sensor fingerprints. 
For example the \gls{prnu} pattern is commonly used in image/video device identification techniques \cite{marra2017blind,mandelli2020cnn, iuliani2019hybrid,mandelli2020facing}.
Since \gls{prnu} patterns can capture the heterogeneity of the sensor response caused by imperfections in the sensor manufacturing process, such sensor fingerprints can attribute a video sequence to one source device uniquely.
Despite being powerful in many forensics tasks, the \gls{prnu} patterns can be difficult to obtain.
The estimation of \gls{prnu} patterns usually requires many samples taken by the device under analysis \cite{lukas2006digital}.
Obtaining the \gls{prnu} patterns from video sequences is more challenging due to the existence of video compression and video stabilization \cite{mandelli2018blind}.
These challenges limit the application of \gls{prnu} patterns in \gls{vcdi} tasks.

H.264 is one of the most popular video compression techniques \cite{iso:avc,richardson2011h264}.
It offers a wide variety of compression configurations to balance data rate, distortion, and computational complexity.
Each configuration will induce a distinct encoding response to the input video sequence.
Even for the same compression configuration, the behavior of H.264 encoder implementations used by different device manufacturers may not be identical. 
Note that the H.264 compression standard only standardizes the decoding but not the encoding \cite{iso:avc,richardson2011h264}.
In this paper we shall refer to the ``encoding pattern'' of a H.264 compressed video sequence as the parameters inserted into the compressed bitstream by the encoder and used by the decoder to reconstruct the video sequence.
We will show in this paper that these encoding patterns along with the video content can be used to determine the source device of an H.264 video sequence.
Since the H.264 encoding pattern tends to be the same for a specific video camera model or firmware version, using it for \gls{vcdi} tasks may only result in a model-level or firmware-level match, which is coarser compared to techniques using video camera sensor fingerprints.
However, these H.264 encoding patterns are closely related to the compressed digital video and are less affected by operations that alter the sensor fingerprints such as video stabilization. 
Compared to metadata forensic methods such as \cite{xiang2021forensic,lopez2020digital, altinisik2022camera}, video forensics methods based on using the video content and encoding parameters can still work even if the metadata information is modified (e.g., when an MP4 video file is converted to an AVI video file without transcoding the video stream).
Our proposed method only requires two H.264 \glspl{gop} to make decisions, which allows our approach to work with corrupted or fragmented H.264 data.

In this paper, we focus on open-set \gls{vcdv}, which is related to \gls{vcdi}.
Open-set classification is the problem of handling classes that are not contained in the training dataset.
Traditional classification approaches assume that only known classes appear in the testing environment  \cite{Scheirer_2013_TPAMI}.
In \gls{vcdi}, the video forensics method is asked to identify which device was used to capture the video sequence.
In \gls{vcdv}, the video forensics method is asked to determine if two  video sequences are captured by the same camera model \cite{mayer2020open,yang2021fast,altinisik2021source}.
With \gls{vcdv} methods, \gls{vcdi} can be achieved by obtaining a video sequence from a known device and then determining if the video sequence under analysis is captured by the same device.
Since \gls{vcdi} methods attribute a given video sequence to a specific source device, these methods require prior knowledge about the devices.
As a result, \gls{vcdi} methods are often constrained to closed-set problems, where the video sequence under test comes from a device that is already known to the video forensics method \cite{mayer2020open}.
This tends to have limitations in practice, as forensic investigators are more likely to deal with open-set problems where the video camera model has not been encountered by the method before.
Because \gls{vcdv} methods only analyze if the two video sequences are from the same device, it is possible for these methods to work in open-set scenarios where the devices have never been seen by the \gls{vcdv} methods before \cite{mayer2020open}.
We tested the open-set performance of our proposed \gls{vcdv} technique and verified that it had good performance evaluation metrics for unseen device models.
The ability to attain open-set \gls{vcdv} allows our method to be used in a wider range of forensic investigations.

The rest of the paper is organized as follows.
In \cref{sec:background}, we show the related work and provide the background knowledge about the H.264 video compression and the machine learning model used in our approach.
In \cref{sec:proposed_method}, we describe the details of our proposed \gls{vcdv} method.
In \cref{sec:experiments_and_results}, we discuss the details of our experiments and present the results.
\cref{sec:conclusion} concludes the paper and gives insights on open problems and future challenges.

\section{Background}\label{sec:background}

\glsreset{vcdi}
\glsreset{vcdv}
\glsreset{gop}

In this section we show existing work related to \gls{vcdi}, \gls{vcdv}, and H.264-based video forensics.
Then, we briefly introduce H.264 video compression and transformer neural networks, which are two important concepts for understanding the mechanism of our proposed open-set \gls{vcdv} method.

\subsection{Related Work}

\glsreset{vcdi}
\glsreset{vcdv}
\glsreset{gop}

\gls{vcdi} is an important topic in video forensics.
In \cite{su2009asource}, the authors used the statistics of motion vectors from video codecs for \gls{vcdi}.
Yahaya \etal\ \cite{yahaya2012advanced} used conditional probability features to achieve \gls{vcdi}.
In \cite{xiang2021forensic,lopez2020digital, altinisik2022camera}, the authors used metadata information stored in video container formats for \gls{vcdi}.
Most existing work on \gls{vcdi} are based on camera sensor noise fingerprints, which was first studied in \cite{kurosawa1999ccd}.
In the seminal work proposed by Chen \etal\ \cite{chen2007source}, \gls{prnu} analysis was first  used for \gls{vcdi} tasks.
In \cite{villalba2016identification,altinisik2020mitigation,mandelli2020facing,yang2021fast,ferrara2022prnu}, the authors devised various strategies to improve the performance of \gls{prnu} analysis of video, such as selecting key frames, counteracting the effects of video stabilization, using the characteristics of the video codecs, and weighting frames in terms of compression quality.
Beyond \gls{prnu} patterns, \cite{hosler2019avideo,verdoliva2019extracting,timmerman2020video} used deep neural networks to extract features from decoded video frames, which can be used in \gls{vcdi} tasks.
Some \gls{vcdi} techniques use multimodal data to improve the performance.
Iuliani \etal\ \cite{iuliani2019hybrid} combined image and video data from the same sensor for better \gls{vcdi} results.
Dal Cortivo \etal\ \cite{dal2021cnn} proposed a \gls{vcdi} approach based on video and audio information.

As described in \cref{sec::introduction}, \gls{vcdv} is a concept derived from \gls{vcdi}.
It is also known as Video Device Verification.
In \cite{yang2021fast,altinisik2021source}, the authors improved \gls{prnu} analysis for \gls{vcdv}.
Mayer \etal\ \cite{mayer2020open} addressed the problem of open-set \gls{vcdv} using a deep neural network to extract features from decoded video frames.

There have been a number of video forensics approaches that use the characteristics of H.264 video compression.
In \cite{vazquez2012detection,xu2015relocated,bestagini2016codec,he2017frame,yao2017detection,yao2020double,vazquez2020video,mahfoudi2022statistical}, the authors used information from the H.264 codec to determine if an H.264 video is double compressed.
Verde \etal\ \cite{verde2018video} used deep neural networks to extract H.264 codec information for video manipulation localization.
We believe our proposed technique is the first to use H.264 codec information for open-set \gls{vcdv} problems.

\subsection{H.264 Video Compression}

The details of the H.264 video compression standard are extremely sophisticated and beyond the scope of this paper.
Therefore, we provide a high-level overview of H.264 video compression that is sufficient for understanding our proposed \gls{vcdv} method.
More details about H.264 compression can be obtained from \cite{iso:avc,richardson2011h264,sullivan2005video}.

One important concept that is a part of nearly all video compression standards is the use of both spatial and temporal redundancy in a video sequence to reduce its data rate, particularly the fact that consecutive frames in a small temporal interval can be greatly correlated.
The H.264 encoder examines the video frames in a structure known as \gls{gop}, which is a sequence of consecutive frames.
Due to the high temporal correlation, the frames in a \gls{gop} can be compressed using motion compensation \cite{richardson2011h264}. 
The frames in a \gls{gop} are divided into I-, P-, and B-frames that are used for the motion compensation.
Typically the first frame in a \gls{gop} is an I-frame that acts as the reference frame for other frames in the GOP. 
The I-frame is compressed using intra-frame compression similar to JPEG.
The rest of the frames in an H.264 \gls{gop} can be P-frames or B-frames, and are compressed with inter-frame compression and motion compensation using the I-frame or a P-frame as a reference frame \cite{richardson2011h264}.

In I-, P-, and B-frames, H.264 uses $16\times 16$ frame patches known as macroblocks in each video frame.
Each macroblock is associated with a macroblock type, which specifies how the information in the macroblock is compressed.
Macroblock-level compression can done by predicting patterns within the macroblock (i.e., intra-coded macroblocks) or using difference information from a similar macroblock in a motion compensated reference frame (i.e., inter-coded macroblocks) \cite{richardson2011h264}.
This difference is known as the prediction residual, which is more efficient to compress compared to the original video frame.
The pixels in the macroblocks are then transformed and quantized before being entropy coded \cite{richardson2011h264}.
The quantization process is controlled by a \gls{qp}.
As the \gls{qp} increases, the data rate decreases and the quantization distortion increases; as \gls{qp} decreases, the rate-distortion trade-off goes towards the opposite direction \cite{richardson2011h264}.
For inter-coded macroblocks the encoded prediction residuals and the motion vectors from the motion compensation are placed in the compressed bitstream.

H.264  uses the YUV color space, which has one luma (luminance) channel and two chroma (chrominance) channels \cite{richardson2011h264}.
Since the human visual system is more sensitive to luminance than chrominance,  H.264  prioritizes the compression quality of the luma channel over the chroma channels to reduce data rate.
Each color channel has its own \gls{qp} for rate-distortion control \cite{richardson2011h264}.

\subsection{Vision Transformers}

Our proposed \gls{vcdv} method is based on transformer neural networks \cite{vaswani2017attention}.
These networks demonstrated outstanding performance in a wide variety of tasks such as language modeling \cite{brown2020language}, image classification \cite{dosovitskiy2020image}, image segmentation \cite{liu2021swin}, video classification \cite{liu2021video}, audio signal processing \cite{verma2021audio}, and protein structure prediction \cite{jumper2021highly}.

Transformers can be used to process sequence data. 
They possess faster computational speed, higher scalability, and better stability compared to Recurrent Neural Networks such as LSTM \cite{hochreiter1997long} and GRU \cite{cho2014properties, zeyer2019comparison}. 
Transformer networks are made up of transformer layers.
The output of a transformer layer is used as the input to the next transformer layer.
Denote the input to a transformer layer by $\bZ \in \mathbb{R}^{N \times D}$, where $N$ is the sequence length and $D$ is the dimensionality of the vector at each time step. 
Transformer layers estimate the relationship between the vectors at each time step in $\bZ$ based on the \gls{sa} mechanism.
In \gls{sa}, the input $\bZ$ is first linearly projected into three matrices $\bq,\bk,\bv \in \mathbb{R}^{N\times D_h}$ using a matrix $\bU \in \mathbb{R}^{D\times 3 D_h}$ such that $[\smallsp\bq\smallsp|\smallsp\bk\smallsp|\smallsp\bv\smallsp]=\bZ\bU$.
Then, the \gls{sa} of $\bZ$ is computed by $\sa(\bZ) = \operatorname{softmax}(\nicefrac{\bq\bk^T}{\sqrt{D_h}})\bv$.
An extended version of \gls{sa} known as \gls{msa} is used in \cite{vaswani2017attention}, where $h$ different \gls{sa} values (known as ``\gls{msa} heads'') are computed at the same time.
The $h$ different \gls{sa} values are combined to form the result of \gls{msa} using a matrix $\bV \in \mathbb{R}^{D \times D}$ such that $\msa(\bZ) = [\sa_1(\bZ) \mid \cdots \mid \sa_h(\bZ)]\bV$. 
When choosing the number of \gls{msa} heads $h$, it must hold that $h$ divides $D$ and $D_h=\nicefrac{D}{h}$, which ensures the output of the transformer layer $\msa(\bZ)$ has the same dimensionality as the input $\bZ$.
Since a transformer network consists of a series of such layers, its input and output share the same dimensionality.

\gls{vit} \cite{dosovitskiy2020image} uses transformer networks for computer vision tasks.
In order to convert image data into the format accepted by transformers, the \gls{vit} splits the image into $K$ non-overlapping patches of size $P \times P \times C$, where $P$ is the patch size and $C$ is the number of channels. 
The patches are flattened to form the matrix $\bZ' \in \mathbb{R}^{K\times P^2C}$. 
Since $P^2C$ is usually large, the dimensionality of the flattened vectors from each patch is reduced to $D_{\mathrm{ViT}}$ using a linear projection $\bW \in \mathbb{R}^{P^2C \times D_{\mathrm{ViT}}}$ such that $\bZ=\bZ'\bW$.
The original input image can be represented by $\bZ$, which is used as the input to the transformer network.

\section{Proposed Method}\label{sec:proposed_method}

\glsreset{vcdi}
\glsreset{vcdv}
\glsreset{gop}
\glsreset{qp}
\glsreset{msa}

\begin{figure}[htpb]
  \centering
  \includegraphics[width=0.75\linewidth]{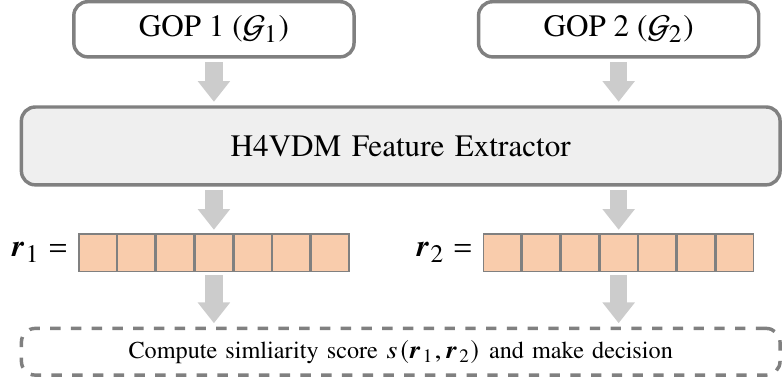}
  \caption{The block diagram of our proposed \appfullname\ method. The details of the \appname\ feature extractor are described in \cref{sec:H4DV-feature-extractor}. Note the input GOP information includes the decoded frame and the coding parameters.}
  \label{fig:main-block-diagram}
\end{figure}

The H.264 \gls{vcdv} problem can be formally defined as follows:
given two \glspl{gop} $\G_1$ and $\G_2$ from two video sequences, determine if the two \gls{gop}s are from the same device. Note the GOP information includes the decoded frame and the coding parameters.
We propose an \appfullname\  method for open-set \gls{vcdv}.
The block diagram of \appname\  is shown in \cref{fig:main-block-diagram}. 
This network architecture is commonly used in many open-set digital forensics techniques \cite{mayer2020open, mandelli2020cnn, mayer2020forensic}.
The input to \appname\  is two H.264 \gls{gop}s $\G_1$ and $\G_2$.
They are then passed to the \appname\  feature extractor separately.

The \appname\  feature extractor is the key component of our proposed method, which is described extensively in \cref{sec:H4DV-feature-extractor}.
It extracts important information from an H.264 \gls{gop} $\G$ and expresses this information in a $D_{\br}$-dimensional vector representation $\br$ known as the \gls{gop} feature vector.
Let $\br_1$ and $\br_2$ be the corresponding \gls{gop} feature vectors of $\G_1$ and $\G_2$, respectively.
We compute a similarity score $s(\br_1, \br_2)$ between the two \gls{gop} feature vectors.
The similarity score can be used for classification: a higher similarity score indicates that $\G_2$ is more likely to be captured by the same device as $\G_1$.
The similarity score used in our proposed method is described in \cref{sec:sim-score}.

\subsection{\appname\  Feature Extractor}\label{sec:H4DV-feature-extractor}

\glsreset{vit}

\begin{figure*}[htpb]
  \centering
  \includegraphics[width=0.95\linewidth]{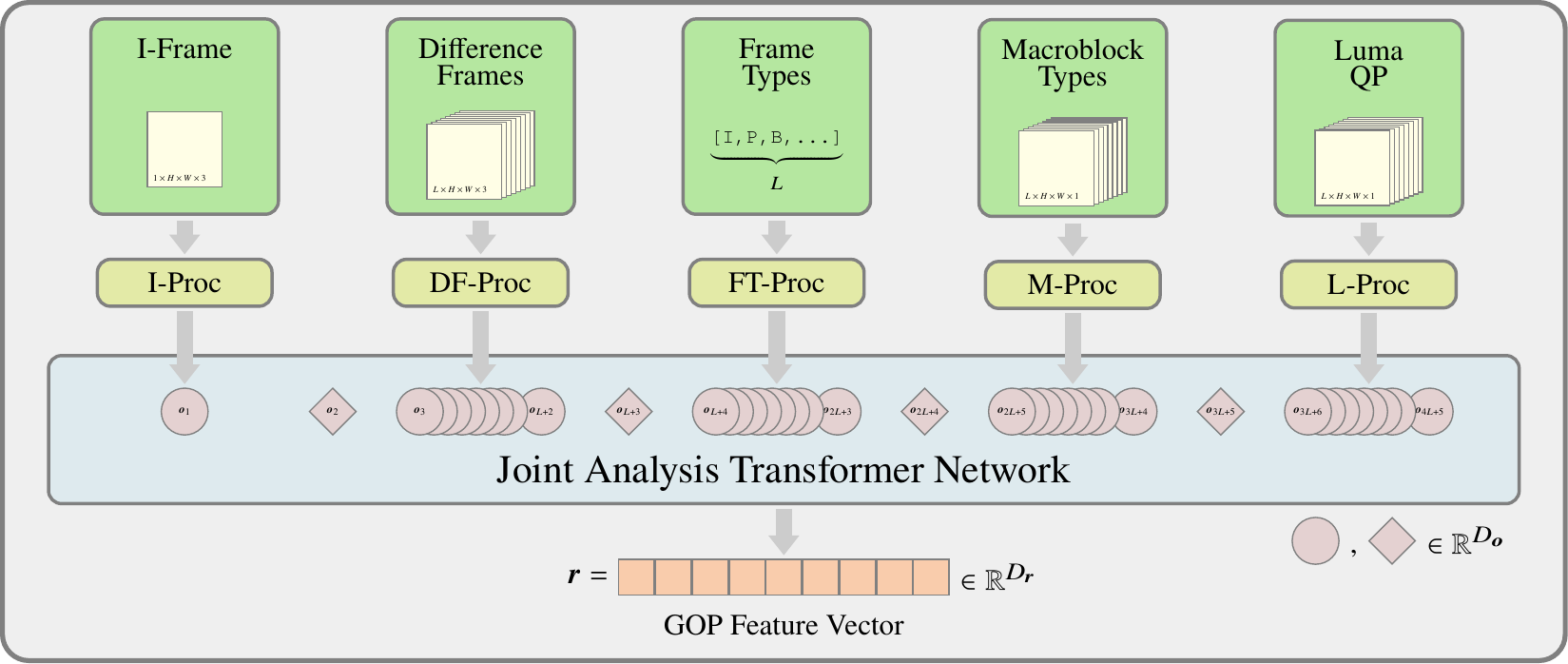}
  \caption{The block diagram of the \appname\  feature extractor.}
  \label{fig:H4DV-feature-extractor}
\end{figure*}

The block diagram of the \appname\  feature extractor is shown in \cref{fig:H4DV-feature-extractor}.
The feature extractor computes \gls{gop} feature vectors using H.264 \gls{gop}s containing $L$ frames, where the size of each frame is $H \times W$.
If a \gls{gop} is longer than $L$ frames, we extract features for the first $L$ frames only without loss of generality.
If the frame size of a \gls{gop} is larger than $H \times W$, we extract features from an arbitrary $H \times W$ region in the \gls{gop} for our analysis.
Feature extraction is not possible when the length of \gls{gop} is less than $L$ or when the \gls{gop} frame size is smaller than $H\times W$.
Therefore, it is important to set $L$, $H$, and $W$ appropriately so that the \appname\  feature extractor can process \gls{gop}s from a wide range of H.264 video sequences.

The \appname\ feature extractor uses five types of data from an H.264 \gls{gop} to generate \gls{gop} feature vectors, i.e., the I-frame, the frame differences, the frame types, the macroblock types and the luma QPs. 
Each type of data is first processed by a specific approach to generate an intermediate output 
denoted by $\bt \in \mathbb{R}^{D_{\bt}}$. 
The five processing methods for each data type are known as the I-, DF-, FT-, M-, and L-Proc.
{
Many of these processing methods use \glspl{vit} to process image-like information.
In total, we use two \gls{vit} architectures in our proposed method, i.e., ViT-1 and ViT-2 (as shown in \cref{tab:vit-architecture}).
ViT-1 is a larger network that processes more complicated data such as I-frames and residual frames.
ViT-2 is a smaller network that processes simpler data such as macroblock types and luma \glspl{qp}.
The details of each processing method are described as below.
}

\paragraph{I-Proc.} This processing step extracts feature from the decoded I-frame in the \gls{gop}. 
The I-Proc uses the ViT-1 architecture in \cref{tab:vit-architecture}.
For each \gls{gop}, the input to the I-Proc is an $H \times W \times 3$ vector consisting of the RGB pixel data from the I-frame.
The output of the I-Proc is $\bt_1$.

\paragraph{DF-Proc.} This processing step extracts features from differences of the decoded frames in sequence.
The differences we compute are between the decoded frames and the decoded I-frame in the \gls{gop} (including the difference between the I-frame and itself, which is all zeros).
Using difference frames can better enable the DF-Proc to learn about the characteristics of H.264 compression.
The DF-Proc also uses the ViT-1 architecture in \cref{tab:vit-architecture}.
For each \gls{gop}, the input to the DF-Proc is $L$ vectors of dimension $H \times W \times 3$, where each vector is the RGB pixel difference.
The $L$ frame difference vectors are processed one by one to generate $L$ outputs $\bt_3, \ldots, \bt_{L+2}$.

\paragraph{FT-Proc.} This step converts frames types into $\dt$ dimensional vectors. 
For each input \gls{gop}, the frame types are a sequence of $L$ positive integers, each of which represents a valid frame type in H.264 (i.e., I, P, B).
The FT-Proc projects each integer to a $\dt$-dimensional real-valued vector with the widely used embedding technique \cite{mikolov2013efficient}.
In the original integer representation, each integer is an index for a concept (e.g., frame types).
The distance between two integer indices is not meaningful.
By converting the integer indices into real-valued vectors using a projection learned in the training phase, the vector representation of similar concepts can have smaller distances, which makes learning easier for the machine learning method.
The final outputs are $L$ vectors $\bt_{L+4}, \ldots, \bt_{2L+3}$.

\paragraph{M-Proc.} Here we extract features from the macroblock types.
In the original H.264 data stream, a frame is subdivided into macroblocks of size $16 \times 16$.
Each macroblock in the frame can be compressed with different methods.
The compression method used in a macroblock is stored as an integer known as the macroblock type \cite{vazquez2020video}.
This macroblock type information is converted into a $H \times W \times 1$ vector by ``unpacking'' the macroblocks.
That is, every pixel in the frame is associated with a macroblock type integer inherited from the macroblock it belongs to.
We use the embedding technique to project each macrotype integer into a three dimensional real-valued vector, which is then processed by a ViT-2 network described in \cref{tab:vit-architecture}.
For the $L$ frames in the \gls{gop}, the output is $L$ vectors $\bt_{2L+5}, \ldots, \bt_{3L+4}$.

\paragraph{L-Proc.} This step extracts features from the luma \gls{qp}s.
The input is an $H \times W \times 1$ vector, where each element is an integer ranging from 0 to 51, i.e., the luma QPs.
As the luma QP increases, the H.264 quantization procedure discards more details in the luma channel in exchange for lower data rate \cite{valenzise2010estimating}.
Due to the ordered nature of luma QPs, we process them directly using a ViT-2 network in \cref{tab:vit-architecture}.
For the $L$ frames in the \gls{gop}, the output is $L$ vectors $\bt_{3L+6}, \ldots, \bt_{4L+5}$.

\begin{table}
  \centering
  \caption{The hyperparameters of the vision transformers (ViT) \cite{dosovitskiy2020image} used in \appname, as discussed in \cref{sec:model_selection}. 
  }
  \small
  \begin{tabular}{lcc}
    \toprule
    \multicolumn{1}{c}{Hyperparameters} & ViT-1 & ViT-2\\ \midrule
    depth &  8 & 4\\
    projection dimension & $D_{\mathrm{ViT}1}$ & $D_{\mathrm{ViT}2}$\\
    number of MSA heads & 8 & 4\\
    output dimension & $\dt$ & $\dt$\\ 
    patch size & 16 & 16\\
    \bottomrule
\end{tabular}
  \label{tab:vit-architecture}
\end{table}

\medskip
The intermediate outputs from the five processing networks contain important information about different data types. 
Similar to \cite{sun2019videobert}, we insert special vectors (i.e., $\bt_2$, $\bt_{L+3}$, $\bt_{2L+4}$, $\bt_{3L+5}$) to combine the information acquired from various data types.
These special vectors can be updated during training.
In total, there are $4L+5$ intermediate output vectors, which are used as the input to the joint analysis network. 
This network is an 8-layer transformer network \cite{vaswani2017attention}.
The output of the joint analysis network is linearly projected to a vector $\br \in \mathbb{R}^{\dr}$. 
The vector $\br$ is the output of the \appname\  feature extractor (i.e., the \gls{gop} feature vector).

Based on the five data types in the input \gls{gop}, the \appname\  feature extractor characterizes macroblock type selection, luma \gls{qp} selection, and other patterns that are specific to the video capturing device.
This information is contained in a $\dr$-dimensional \gls{gop} feature vector. 
By comparing the similarity score between the corresponding \gls{gop} feature vectors from two video sequences we can determine if the two video sequences were captured by the same device.
Since \appname\ only requires two H.264 \gls{gop}s to make decisions, it is able to work in scenarios where data from a test device is scarce, e.g., when the H.264 video sequence is corrupted and only a few \gls{gop}s can be recovered.

\subsection{Similarity Score and Loss Function}\label{sec:sim-score}

\begin{figure}[htpb]
  \centering
  \includesvg[width=0.7\linewidth, pretex=\small]{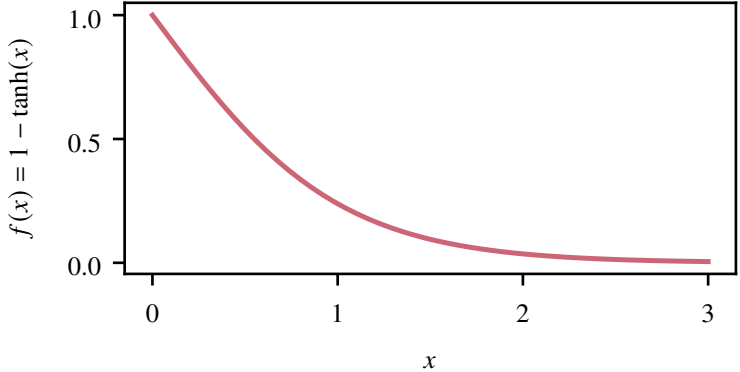}
  \caption{The similarity score function.}
  \label{fig:sim_score}
\end{figure}

The similarity score between two \gls{gop} feature vectors $\br_1$ and $\br_2$ is a real number in the range $[0,1]$, where $1$ indicates the two vectors are the most similar and $0$ indicates the two vectors are the most dissimilar.
We compute the similarity score using the following function
\begin{align}\label{eqn:sim_score}
  s(\br_1, \br_2) = 1 - \tanh\left(\lVert \br_1 - \br_2 \rVert_2\right),
\end{align}
where $\lVert \cdot \rVert_2$ denotes $L_2$-norm. 
The function $f(x)=1-\tanh(x)$ for $x \geq 0$ is shown in \cref{fig:sim_score}.
When $\br_1$ and $\br_2$ are more similar, $\lVert \br_1 - \br_2 \rVert_2 \rightarrow 0$, which indicates $s(\br_1, \br_2) \rightarrow 1$.
Conversely, $s(\br_1, \br_2) \rightarrow 0$ when $\br_1$ and $\br_2$ are more dissimilar.

We use binary cross-entropy loss and the similarity score (\cref{eqn:sim_score}) to compute the loss of the \appname\  method during training.
Suppose the ground truth label of the \gls{gop} feature vector pair $(\br_1, \br_2)$ is given by $y$, where $y=1$ indicates the two \gls{gop} features are from the same video capturing device and $y=0$ indicates the opposite.
The loss of the pair is
\begin{equation}
  \begin{aligned}
    \ell(\br_1, \br_2, y) = &y \log\left[s(\br_1, \br_2) \right]  + (1-y) \log\left[1-s(\br_1, \br_2) \right].
  \end{aligned}
\end{equation}

\section{Experiments and Results}\label{sec:experiments_and_results}

We describe our experiments in this section including the datasets used for training and testing.
More details about \appname\  such as hyperparameter selection and training strategy are also discussed.
Finally, we present the results from the experiments.
We used \gls{auc} score \cite{huang2005using}, $F_1$-score \cite{hripcsak2005agreement}, and accuracy score to evaluate the performance of \appname.
Throughout the experiments, we selected the dimensionality of the \gls{gop} feature vectors to be 1024, i.e., $\dr=1024$.

\subsection{Dataset Generation}

The datasets used in our experiments are generated from the VISION dataset \cite{shullani2017vision}.
The VISION dataset contains 648 H.264 video sequences from 35 different video capturing devices.
The list of devices in VISION is shown in \cref{tab:vision_devices}.

\begin{table}[htpb]
  \centering
  \caption{The list of devices contained in the VISION dataset \cite{shullani2017vision}.}
  \bgroup
  \small
  \catcode`_=12
    \begin{tabular}{clcl}
\toprule
\makecell{Device\\ ID} & \makecell{Device\\ Name} & \makecell{Device\\ ID} & \makecell{Device\\ Name}\\ \midrule
1 & Samsung_GalaxyS3Mini & 19 & Apple_iPhone6Plus (iOS 10.2.1)\\ 
2 & Apple_iPhone4s (iOS 7.1.2) & 20 & Apple_iPadMini (iOS 8.4)\\ 
3 & Huawei_P9 & 21 & Wiko_Ridge4G\\ 
4 & LG_D290 & 22 & Samsung_GalaxyTrendPlus\\ 
5 & Apple_iPhone5c (iOS 10.2.1) & 23 & Asus_Zenfone2Laser\\ 
6 & Apple_iPhone6 (iOS 8.4) & 24 & Xiaomi_RedmiNote3\\ 
7 & Lenovo_P70A & 25 & OnePlus_A3000\\ 
8 & Samsung_GalaxyTab3 & 26 & Samsung_GalaxyS3Mini\\ 
9 & Apple_iPhone4 (iOS 7.1.2) & 27 & Samsung_GalaxyS5\\ 
10 & Apple_iPhone4s (iOS 8.4.1) & 28 & Huawei_P8\\ 
11 & Samsung_GalaxyS3 & 29 & Apple_iPhone5 (iOS 9.3.3)\\ 
12 & Sony_XperiaZ1Compact & 30 & Huawei_Honor5c\\ 
13 & Apple_iPad2 (iOS 7.1.1) & 31 & Samsung_GalaxyS4Mini\\ 
14 & Apple_iPhone5c (iOS 7.0.3) & 32 & OnePlus_A3003\\ 
15 & Apple_iPhone6 (iOS 10.1.1) & 33 & Huawei_Ascend\\ 
16 & Huawei_P9Lite & 34 & Apple_iPhone5 (iOS 8.3)\\ 
17 & Microsoft_Lumia640LTE & 35 & Samsung_GalaxyTabA\\ 
18 & Apple_iPhone5c (iOS 8.4.1)\\ \bottomrule
\end{tabular}
  \egroup
  \label{tab:vision_devices}
\end{table}

We decoded the video frames and extracted the \gls{gop}s from video sequences in VISION using a customized version of the \texttt{openh264}\footnotemark\ H.264 decoder, which allows us access to the GOP information.
\footnotetext{\url{https://github.com/cisco/openh264}}
In our analysis, we selected the length of \gls{gop} $L=8$. That is, our method can analyze H.264 \gls{gop} with length greater than or equal to 8.
The height ($H$) and width ($W$) of the frames were set to $H=W=224$, which is a common size used by popular image or video processing techniques such as \cite{he2015deep,dosovitskiy2020image,liu2021swin,liu2021video}.
With this choice of $H$ and $W$, the frame sizes of all video sequences in the VISION dataset are larger than $(H,W)$.
Therefore, we cropped a $(H,W)$ region from the center of the frames for analysis.
We chose to crop the region at the center because the video content at the center is more likely to change compared to those from the edges or corners.
From each video sequence in the VISION dataset, we randomly sampled 15 \gls{gop}s whose length is greater than or equal to $L$.
When the length of a sampled \gls{gop} is greater than $L$, only the first $L$ frames in the \gls{gop} were used.

We constructed data from the VISION dataset to train and test our method as follows:
\begin{enumerate}[itemsep=1pt]
  \item Provide a set of device indices $\mathcal{S}$, selected from \cref{tab:vision_devices}.
  \item Select all pairs of device indices $(i,j) \in \mathcal{S} \otimes \mathcal{S}$, where $\otimes$ denotes Cartesian product.
  Denote the set of \gls{gop}s from device $i$ and device $j$ by $\mathcal{A}_i$ and $\mathcal{A}_j$, respectively. For each $(i,j)$ do the following:
  \begin{enumerate}[topsep=0pt]
    \item If $i\neq j$, randomly sample $n_0$ unique \gls{gop} pairs from the set $\mathcal{A}_i \otimes \mathcal{A}_j$ that are not in the dataset. 
    When determining if a \gls{gop} pair is in the dataset, the \gls{gop} pair is considered to be unordered. 
    That is, if one swaps the first and the second element, the pair is considered to be the same.
    These $n_0$ pairs are assigned label $0$ and added to the dataset.
    \item If $i=j$, randomly sample $n_1$ unique \gls{gop} pairs from the set $\mathcal{A}_i \otimes \mathcal{A}_j$ that are not in the dataset. 
    When determining if a \gls{gop} pair is in the dataset, the \gls{gop} pair is considered to be unordered.
    These $n_1$ pairs are assigned label $1$ and added to the dataset.
  \end{enumerate}
\end{enumerate}

In our experiments, we chose $n_0=15$ and $n_1=120$.
To better evaluate the performance of H4VDM, we constructed 7 datasets, where each dataset contained a training set and a testing set.
For brevity, we refer to these datasets as D1, D2, \ldots, D7.
In each dataset, the set of all device IDs in VISION are split into two disjoint sets $\mathcal{S}_1$ and $\mathcal{S}_2$.
They are passed to the dataset construction step to generate the training set and the testing set, respectively.
For dataset D1--D4, $\mathcal{S}_2$ contained half of the device IDs that were randomly selected.
For dataset D5--D7, the device IDs are split into three disjoint sets and used as $\mathcal{S}_2$ of each dataset.
The details of each dataset are shown in \cref{tab:dataset_fold_detail}.
For the testing set of each dataset, we uniformly sampled 40 percent of the GOP pairs after dataset generation to reduce its size.
In each dataset, we uniformly removed $\nicefrac{1}{8}$ of the \gls{gop} pairs from the testing set for validation during training.
The training process is stopped when the open-set validation performance no longer increases.

\begin{table}[htpb]
  \centering
  \caption{
  The details of each dataset.
  \#0 and \#1 indicates the number of \gls{gop} pairs with class 0 (different device) and class 1 (same device), respectively.
  }
  \resizebox{0.95\linewidth}{!}{
  \setlength{\tabcolsep}{1ex}
    \begin{tabular}{*{6}{c}}
    \toprule
    \multirow{2}{*}{Dataset} &
    \multirow{2}{*}{Test Device IDs ($\mathcal{S}_2$)} &
    \multicolumn{2}{c}{Training} &
    \multicolumn{2}{c}{Testing}\\ \cmidrule(lr{1pt}){3-4}\cmidrule(lr{1pt}){5-6}
    & & \#0 & \#1 & \#0 & \#1
    \\ \midrule
    D1 & 
    \makecell{\{1, 2, 4, 5, 6, 14, 17, 18, 19, 21, 22, 23, 27, 28, 30, 32, 35\}}
    & 4590 & 2160 & 816 & 1632
    \\
    D2 & 
    \makecell{\{1, 2, 4, 11, 14, 15, 17, 18, 19, 20, 21, 23, 26, 30, 32, 33, 35\}} 
    & 4590 & 2160 & 816 & 1632
    \\
    D3 & 
    \makecell{\{4, 5, 6, 10, 11, 13, 14, 16, 17, 19, 21, 22, 23, 30, 31, 32, 35\}} 
    & 4590 & 2160 & 816 & 1632
    \\
    D4 & 
    \makecell{\{3, 4, 6, 8, 13, 17, 19, 20, 21, 22, 23, 26, 29, 30, 31, 32, 34\}} 
    & 4590 & 2160 & 816 & 1632
    \\ \midrule
    D5 & \makecell{\{1, 4, 7, 10, 13, 16, 19, 22, 25, 28, 31, 34\}}
    & 7590 &  2760 & 792 & 576
    \\
    D6 & \makecell{\{2, 5, 8, 11, 14, 17, 20, 23, 26, 29, 32, 35\}}
    & 7590 &  2760 & 792 & 576
    \\
    D7 & \makecell{\{3, 6, 9, 12, 15, 18, 21, 24, 27, 30, 33\}}
    & 8280 & 2880 & 660 & 528
    \\
    \bottomrule
\end{tabular}
  }
  \label{tab:dataset_fold_detail}
\end{table}

We selected $L=8$ in our experiments based on several factors.
Increasing $L$ is likely to improve the quality of \gls{gop} feature extraction, because the \appname\  feature extractor is exposed to more information.
However, it will increase the complexity of the model and make the training process more computationally intense.
Since our method requires the length of the \gls{gop} under analysis to be at least $L$, a larger $L$ value can also reduce the number of valid \gls{gop} candidates in a video sequence.
Note that it is computationally expensive to acquire the decoded frames and \gls{gop} information from the H.264 bitstream. 
In training, since the training data will be used repeatedly in each training epoch, it is more efficient to store the decoded frames and \gls{gop} information on disk.
Therefore, another factor that affects the choice of $L$ is the storage size of the uncompressed \gls{gop}s and decoded frames.
With the current dataset configurations, each dataset (training set and testing set) requires approximately 176GB of disk space.
Given such an enormous dataset size, the training process is heavily bounded by the I/O performance of the storage devices, which is much slower compared to other hardware components (e.g., GPUs).
In the worst case, the training time can increase linearly with respect to $L$, which is costly when $L$ is large.
We used $L=8$ as a balance among model performance, model complexity, the usability of our method, training complexity, and training speed.
Typical training times were 3-6 hours using six 48GB GPUs.

\subsection{Parameter Initialization and Training}

From \cref{fig:sim_score} it can be seen that the gradients of the similarity score function $f(x)$ will saturate when $x$ is large, which can significantly reduce the efficiency of gradient-based learning.
Therefore, when initializing the model parameters, we limited the scale of the weights in the output layer to be within $[-0.002, 0.002]$. 
This reduced the initial scale of $\lVert \br_1 - \br_2 \rVert_2$, which can facilitate training.

We trained the \appname\  method using the Adam optimizer \cite{kingma2014adam} with a minibatch size of 72.
We used 5 warm-up epochs, where the learning rate linearly increased from $0$ to $8 \times 10^{-6}$.
After warm-up epochs, we trained the method using an initial learning rate of $8 \times 10^{-6}$.
An exponential learning rate decay was used with a decay factor of $0.97$.
During training, we monitored the \gls{auc} score on the validation set.
The training was stopped when the validation \gls{auc} score no longer increased.

\subsection{Model Size Selection}\label{sec:model_selection}

We tuned the hyperparameters $D_{\mathrm{ViT1}}$ and $\dt$ to control the size of the model.
We constructed 3 models of various sizes (i.e., S (small), B (baseline), and L (large)).
For each model, we computed the best \gls{auc} score on the testing set of Dataset D1.
The results are shown in \cref{tab:hyperparameter-auc}.
Since the \appname-B model achieved the best \gls{auc} score, further experiments were conducted with this model.

\begin{table}[htpb]
  \centering
  \caption{The best AUC scores of various models on the testing set of Dataset D1.}
  \begin{tabular}{*{6}{c}}
    \toprule
    Model & $D_{\mathrm{ViT}1}$ & $D_{\mathrm{ViT}2}$ & $\dt$ & Parameters & AUC\\ \midrule
    \appname-S & 192 & 64 & 192 & 48.93M &  79.8\\
    \appname-B & 256 & 64 & 256 & 80.10M & 80.2\\
    \appname-L & 320 & 64 & 320 & 118.95M & 69.4\\ \bottomrule
\end{tabular}
  \label{tab:hyperparameter-auc}
\end{table}

\subsection{Results}\label{sec:results}

\begin{figure*}[htpb]
  \centering
  \newcommand{\ntxt}[1]{\raisebox{1.5pt}{\scalebox{0.4}{\fontsize{9}{2}\selectfont\color{cyan}#1}}}
  \includesvg[height=0.95\textheight, pretex=\scriptsize]{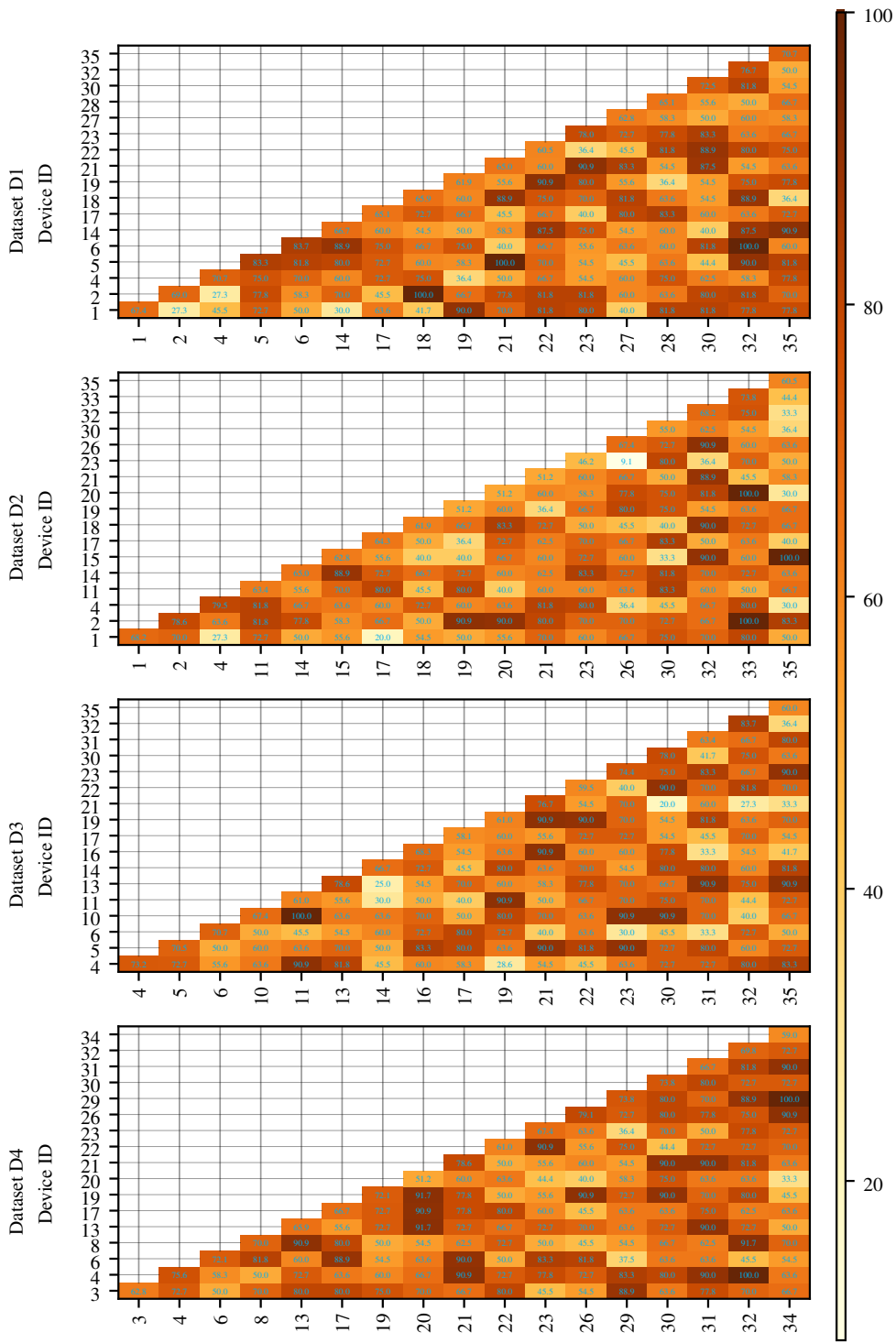}
  \caption{The testing accuracy score matrix of device index pairs on datasets D1--D4.}
  \label{fig:acc_matrix}
\end{figure*}

\paragraph{Results on Datasets D1--D4.} 
In \cref{table:perf_cv}, we show the performance of the \appname-B model on datasets D1--D4.
When computing the $F_1$-score, we selected the threshold such that the sum of \gls{tpr} and \gls{tnr} is maximized.
Our method achieved an overall $F_1$-score of 67.2 and an average \gls{auc} score of 77.4 on datasets D1--D4.
Overall, for class 0 (\gls{gop} pairs from different devices) the precision is high, which means most retrieved \gls{gop} pairs are relevant.
For class 1 (\gls{gop} pairs from the same device) the recall is high, which means most relevant \gls{gop} pairs are retrieved.
In \cref{fig:acc_matrix}, we show the accuracy score matrix of device index pairs of each dataset in testing.
From this figure, it can be seen that H4VDM can match devices at a firmware level.
For example, device 29 and 34 in dataset D4 are the same device (Apple iPhone5) with different operating system versions.
H4VDM was able to distinguish them with high accuracy.

\begin{table}[htpb]
  \centering
  \caption{The testing performance of the \appname-B model on datasets D1--D4.}
  \setlength{\tabcolsep}{1ex}
  \small
    \begin{tabular}{*{11}{c}}
    \toprule
    \multirow{2}{*}{Dataset} &
    \multicolumn{3}{c}{Class 0}&
    \multicolumn{3}{c}{Class 1}&
    \multicolumn{4}{c}{All Classes}
    \\ \cmidrule(lr{3pt}){2-4}\cmidrule(lr{3pt}){5-7}\cmidrule(lr{3pt}){8-11}
    & 
    Pre. &
    Rec. &
    $F_1$ &
    Pre. &
    Rec. &
    $F_1$ &
    Pre. &
    Rec. &
    $F_1$ &
    AUC\\ \midrule
    D1 &
    95.9 &
    53.6 &
    68.8 &
    50.6 &
    95.4 &
    66.1 &
    80.8 &
    67.5 & 
    67.9 &
    80.2
    \\
    D2 &
    91.7 &
    49.7 &
    64.5 &
    47.6 &
    91.1 &
    62.5 &
    77.0 &
    63.5 &
    63.8 &
    74.1
    \\
    D3 &
    89.9 &
    55.4 &
    68.5 &
    49.4 &
    87.5 &
    63.2 &
    76.4 &
    66.1 &
    66.8 &
    78.0
    \\ 
    D4 &
    86.0 &
    64.0 &
    73.4 &
    52.3 &
    79.1 &
    63.0 &
    74.8 &
    69.0 &
    69.9 &
    77.3
    \\ \midrule
    Overall &
    90.5 &
    55.7 &
    68.9 &
    49.9 &
    88.3 &
    63.7 &
    76.9 &
    66.5 &
    67.2 &
    77.4
    \\
     \bottomrule\\[-0.8em]
    \multicolumn{11}{c}{
        Pre.=Precision \hspace*{3em} Rec.=Recall
    }
\end{tabular}
%\hfill Pre.=Precision \hfill Rec.=Recall\hfill

  \label{table:perf_cv}
\end{table}

\paragraph{Results on Datasets D5--D7.}
From the results of H4VDM on datasets D1--D4 (\cref{fig:acc_matrix}), it can be seen that the performance of H4VDM was low for specific device pairs.
This may be caused by the fact that datasets D1--D4 contain only a small number of devices in the training set, which makes it difficult for the method to generalize to a broader scope of devices.
To test the performance of H4VDM on datasets with more training devices, we trained and tested \appname\ on datasets D5--D7, whose training sets contain more devices in the VISION dataset.
The testing performance of the \appname-B model on datasets D5--D7 is shown in \cref{table:dataset_more_train_data}.
The accuracy score matrix of device index pairs is shown in \cref{fig:acc_matrix_old}.
On datasets D5--D7, our method achieved an overall $F_1$-score of 78.6 and an average \gls{auc} score of 85.2.
It can be seen that as the number of devices in the training set increases, the performance of H4VDM becomes better.
However, since the number of devices in the testing set decreases, the testing performance can vary significantly depending on the choice of the testing devices.
Overall, the results on datasets D1--D4 show that H4VDM can learn to achieve open-set VDM given a small number of training devices.
The results on datasets D5--D7 show that the performance of H4VDM can improve quickly as more devices are available in training.

\begin{table}[htpb]
  \centering
  \caption{
  The testing performance of the \appname-B model on datasets D5--D7.
  }
  \newcommand{\subfloat}[4]{
      \parbox[t]{#4}{
        \begin{center}
            \textbf{#1} #2\\ \vspace*{1ex}
            #3
        \end{center}
      }
  }
  \centering
    \setlength{\tabcolsep}{1ex}
    \small
    \begin{tabular}{*{11}{c}}
    \toprule
    \multirow{2}{*}{Dataset} &
    \multicolumn{3}{c}{Class 0}&
    \multicolumn{3}{c}{Class 1}&
    \multicolumn{4}{c}{All Classes}
    \\ \cmidrule(lr{3pt}){2-4}\cmidrule(lr{3pt}){5-7}\cmidrule(lr{3pt}){8-11}
    & 
    Pre. &
    Rec. &
    $F_1$ &
    Pre. &
    Rec. &
    $F_1$ &
    Pre. &
    Rec. &
    $F_1$ &
    AUC\\ \midrule
    D5 &
    85.9 &
    74.5 &
    79.8 &
    75.5 &
    86.5 &
    80.6 &
    80.9 &
    80.2 & 
    80.2 &
    87.0
    \\
    D6 &
    96.1 &
    73.4 &
    83.2 &
    76.9 &
    96.7 &
    85.7 &
    86.9 &
    84.5 &
    84.4 &
    90.0
    \\
    D7 &
    94.8 &
    48.8 &
    64.4 &
    65.5 &
    97.3 &
    78.3 &
    80.1 &
    73.0 &
    71.3 &
    78.5
    \\ \midrule
    Overall &
    92.3 &
    65.6 &
    75.8 &
    72.6 &
    93.5 &
    81.5 &
    82.6 &
    79.2 &
    78.6 &
    85.2
    \\
     \bottomrule\\[-0.8em]
    \multicolumn{11}{c}{
        Pre.=Precision \hspace*{3em} Rec.=Recall
    }
\end{tabular}
%\hfill Pre.=Precision \hfill Rec.=Recall\hfill

  \label{table:dataset_more_train_data}
\end{table}

\begin{figure*}[htpb]
  \centering
  \newcommand{\ntxt}[1]{\raisebox{1pt}{\scalebox{0.4}{\fontsize{10}{2}\selectfont\color{cyan}#1}}}
  \includesvg[height=0.95\textheight, pretex=\scriptsize]{figure/acc_mat_2.svg}
  \caption{The testing accuracy score matrix of device index pairs on datasets D5--D7.}
  \label{fig:acc_matrix_old}
\end{figure*}

\section{Conclusion}\label{sec:conclusion}

In this paper we proposed an H.264-based open-set \gls{vcdv} method known as \appname.
\appname\ uses transformer neural networks to process five types of data from the H.264 decoded frames and the \gls{gop}s.
We trained and tested the \appname-B model on datasets generated from the VISION dataset \cite{shullani2017vision}.
The experimental results showed that \appname\ demonstrated good \gls{vcdv} performance on unseen devices.

Despite the good performance, \appname\ has still room for improvement. 
When selecting hyperparameters, we greedily used the model that had the best performance on Dataset D1.
The selected model may not have the best overall performance across all datasets.
Some important H.264 codec information such as motion vectors and true prediction residuals were not used in \appname.
The small number of devices in the VISION dataset also limited the performance evaluation of H4VDM.
On datasets D1--D4, less dataset bias is introduced in training/testing split, but the performance of H4VDM is relatively low due to small number of devices in the training set.
On datasets D5--D7, the performane of H4VDM is higher, but the influence of dataset bias is stronger due to the small number of devices in the testing set, which resulted in fluctuating testing perforamnce.
A dataset with more devices is required to evaluate the performance of H4VDM more comprehensively.

In future work, we will examine the use of motion vectors and prediction residuals.
We will develop methods that use popular deep learning frameworks efficiently so that the training speed is less constrained by hardware I/O speed.
We will collect video data from more video capturing devices for future video forensics research.
We are investigating other video compression techniques including H.265, H.266, VP9, and AV1.

\bgroup

\egroup

\section*{Acknowledgments}\label{sec:acknowledgment}
This material is based on research sponsored by the Defense Advanced Research Projects Agency (DARPA) and Air Force Research Laboratory (AFRL) under agreement number FA8750-20-2-1004. 
The U.S. Government is authorized to reproduce and distribute reprints for Governmental purposes notwithstanding any copyright notation thereon. The views and conclusions contained herein are those of the authors and should not be interpreted as necessarily representing the official policies or endorsements, either expressed or implied, of DARPA, AFRL or the U.S. Government. Address all correspondence to Edward J. Delp, \texttt{ace@ecn.purdue.edu}.

%\bibliographystyle{splncs04}
%\bibliography{egbib}

\AtNextBibliography{\small}
\printbibliography

\end{document}